\documentclass[11pt]{article}
\usepackage{acl2016}
\usepackage{times}
\usepackage{latexsym}
\usepackage{url}
\usepackage{etoolbox}

\usepackage{graphicx}
\usepackage{url}
\usepackage{amsmath}
\usepackage{amssymb}
\usepackage{subfigure}
\usepackage{color}
\usepackage{caption} 
\usepackage{multirow}
\usepackage{comment}
\usepackage{pdflscape}

\aclfinalcopy 

\title{Towards Abstraction from Extraction: Multiple Timescale Gated Recurrent Unit for Summarization}

\author{Minsoo Kim\\
  School of Electronics Engineering \\
  Kyungpook National University \\
  Daegu, South Korea \\
  {\tt minsoo9574@gmail.com} \\\And
  \bf Moirangthem Dennis Singh \\
 School of Electronics Engineering \\
  Kyungpook National University \\
  Daegu, South Korea \\
  {\tt mdennissingh@gmail.com} \AND
 Minho Lee \\
 School of Electronics Engineering \\
  Kyungpook National University \\
  Daegu, South Korea \\
  {\tt mholee@gmail.com} \\}

\begin{document}

\maketitle

\begin{abstract}
In this work, we introduce temporal hierarchies to the sequence to sequence (seq2seq) model to tackle the problem of abstractive summarization of scientific articles. The proposed Multiple Timescale model of the Gated Recurrent Unit (MTGRU) is implemented in the encoder-decoder setting to better deal with the presence of multiple compositionalities in larger texts. The proposed model is compared to the conventional RNN encoder-decoder, and the results demonstrate that our model trains faster and shows significant performance gains. The results also show that the temporal hierarchies help improve the ability of seq2seq models to capture compositionalities better without the presence of highly complex architectural hierarchies.

\end{abstract}

\section{Introduction and Related Works}
Summarization has been extensively researched over the past several decades. \newcite{Jones:2007} and \newcite{Nenkova:2011} offer excellent overviews of the field. Broadly, summarization methods can be categorized into extractive approaches and abstractive approaches \cite{Hahn:2000}, based on the type of computational task. Extractive summarization is a selection problem, while abstractive summarization requires a deeper semantic and discourse understanding of the text, as well as a novel text generation process. Extractive summarization has been the focus in the past, but abstractive summarization remains a challenge. 

Recently, sequence-to-sequence (seq2seq) recurrent neural networks (RNNs) have seen wide application in a number of tasks. Such RNN encoder-decoders \cite{ChoSMT:2014,Bahdanau:2014} combine a representation learning encoder and a language modeling decoder to perform mappings between two sequences. Similarly, recent works have proposed to cast summarization as a mapping problem between an input sequence and a summary sequence. Recent successes such as \newcite{Namas:2015}; \newcite{Nallapati2016} have shown that the RNN encoder-decoder performs remarkably well in summarizing short text. Such seq2seq approaches offer a fully data-driven solution to both semantic and discourse understanding and text generation.

While seq2seq presents a promising way forward for abstractive summarization, extrapolating the methodology to other tasks, such as the summarization of a scientific article, is not trivial. A number of practical and theoretical concerns arise: 1) We cannot simply train RNN encoder-decoders on entire articles: For the memory capacity of current GPUs, scientific articles are too long to be processed whole via RNNs. 2) Moving from one or two sentences, to several sentences or several paragraphs, introduces additional levels of compositionality and richer discourse structure. How can we improve the conventional RNN encoder-decoder to better capture these? 3) Deep learning approaches depend heavily on good quality, large-scale datasets. Collecting source-summary data pairs is difficult, and datasets are scarce outside of the newswire domain.

In this paper, we present a first, intermediate step towards end-to-end abstractive summarization of scientific articles. Our aim is to extend seq2seq based summarization to larger text with a more complex summarization task. To address each of the issues above, 1) We propose a paragraph-wise summarization system, which is trained via paragraph-salient sentence pairs. We use Term Frequency-Inverse Document Frequency (TF-IDF) \cite{Luhn:1958,JONESIDF:1972} scores to extract a salient sentence from each paragraph. 2) We introduce a novel model, Multiple Timescale Gated Recurrent Unit (MTGRU), which adds a temporal hierarchy component that serves to handle multiple levels of compositionality. This is inspired by an analogous concept of temporal hierarchical organization found in the human brain, and is implemented by modulating different layers of the multilayer RNN with different timescales \cite{MTRNN:2008}. We demonstrate that our model is capable of understanding the semantics of a multi-sentence source text and knowing what is important about it, which is the first necessary step towards abstractive summarization. 3) We build a new dataset of Computer Science (CS) articles from ArXiv.org, extracting their Introductions from the LaTeX source files. The Introductions are decomposed into paragraphs, each paragraph acting as a natural unit of discourse.

Finally, we concatenate the generated summary of each paragraph to create a non-expert summary of the article's Introduction, and evaluate our results against the actual Abstract. We show that our model is capable of summarizing multiple sentences to its most salient part on unseen data, further supporting the larger view of summarization as a seq2seq mapping task. We demonstrate that our MTGRU model satisfies some of the major requirements of an abstractive summarization system. We also report that MTGRU has the capability of reducing training time significantly compared to the conventional RNN encoder-decoder.  

The paper is structured as follows: Section \ref{sec:model} describes the proposed model in detail. In Section \ref{sec:results}, we report the results of our experiments and show the generated summary samples. In Section \ref{sec:discussion} we analyze the results of our model and comment on future work.

\begin{figure}
    \centering
    \includegraphics[width=1\columnwidth]{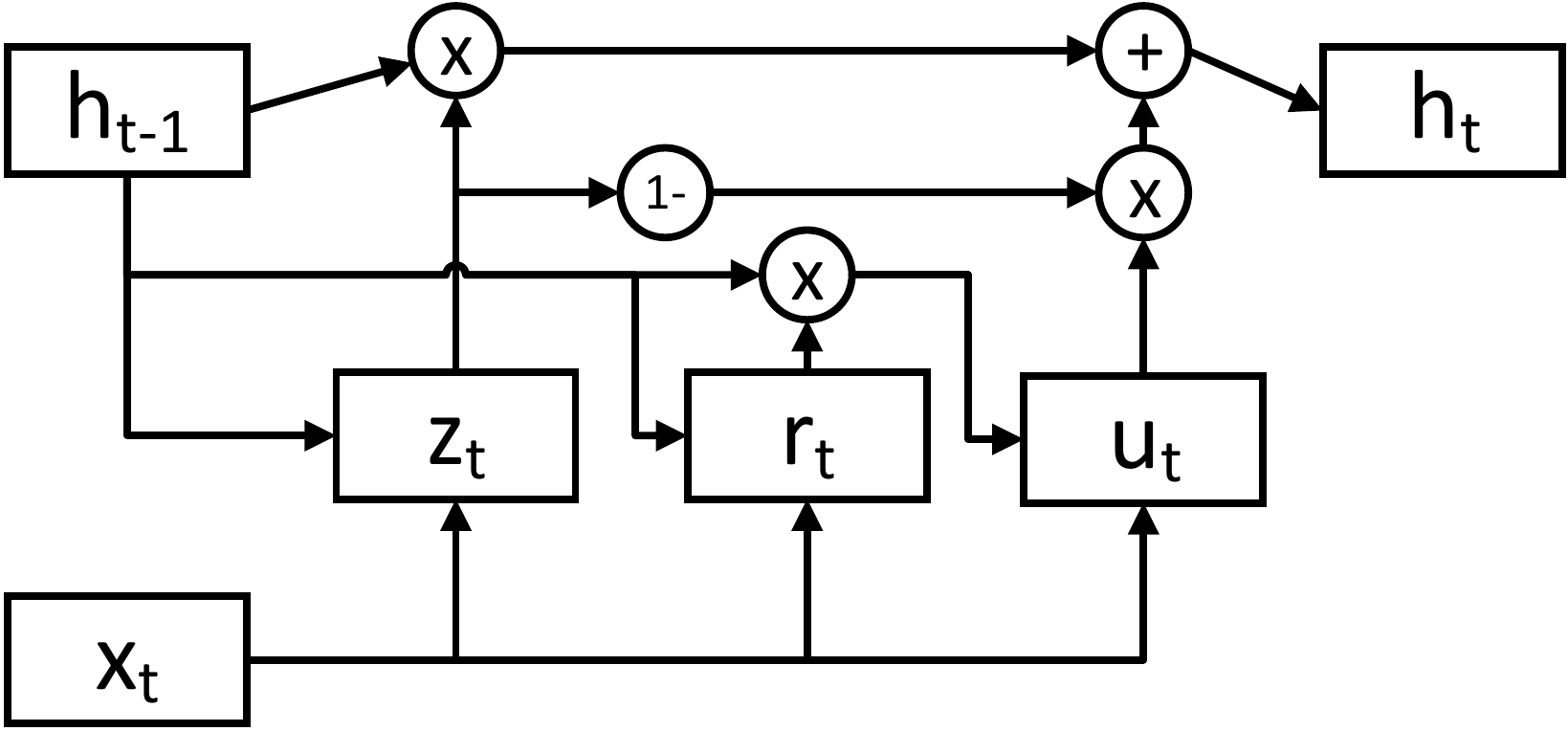}
    \caption{A gated recurrent unit.}
    \label{fig:gru}
    \vspace{-3mm}
\end{figure}

\section{Proposed Model}
\label{sec:model}
In this section we discuss the background related to our model, and describe in detail the newly developed architecture and its application to summarization.

\subsection{Background}
The principle of compositionality defines the meaning conveyed by a linguistic expression as a function of the syntactic combination of its constituent units. In other words, the meaning of a sentence is determined by the way its words are combined with each other. In multi-sentence text, sentence-level compositionality (the way sentences are combined with one another) is an additional function which will add meaning to the overall text. When dealing with such larger texts, compositionality at the sentence and even paragraph levels should be considered, in order to capture the text meaning completely. An approach explored in recent literature is to create dedicated architectures in a hierarchical fashion to capture subsequent levels of compositionality: \newcite{LiHRNN:2015} and \newcite{Nallapati2016} build dedicated word and sentence level RNN architectures to capture compositionality at different levels of text-units, leading to improvements in performance.

However, architectural modifications to the RNN encoder-decoder such as these suffer from the drawback of a major increase in both training time and memory usage. Therefore, we propose an alternative enhancement to the architecture that will improve performance with no such overhead. We draw our inspiration from neuroscience, where it has been shown that functional differentiation occurs naturally in the human brain, giving rise to temporal hierarchies \cite{Meunier2010,Botvinick:2007}. It has been well documented that neurons can hierarchically organize themselves into layers with different adaptation rates to stimuli. The quintessential example of this phenomenon is the auditory system, in which syllable level information in a short time window is integrated into word level information over a longer time window, and so on. Previous works have applied this concept to RNNs in movement tracking \cite{Paine20041291} and speech recognition \cite{Heinrich2012}. 

\subsection{Multiple Timescale Gated Recurrent Unit}
Our proposed Multiple Timescale Gated Recurrent Unit (MTGRU) model applies the temporal hierarchy concept to the problem of seq2seq text summarization, in the framework of the RNN encoder-decoder. Previous works such as \cite{MTRNN:2008}'s Multiple Timescale Recurrent Neural Network (MTRNN) have employed temporal hierarchy in motion prediction. However, MTRNN is prone to the same problems present in the RNN, such as difficulty in capturing long-term dependencies and vanishing gradient problem \cite{hochreiter2001}. Long Short Term Memory network \cite{hochreiter2001} utilizes a complex gating architecture to aid the learning of long-term dependencies and has been shown to perform much better than the RNN in tasks with long-term temporal dependencies such as machine translation \cite{S2S_Ilya:2014}. Gated Recurrent Unit (GRU) \cite{ChoSMT:2014}, which has been proven to be comparable to LSTM \cite{Chung:2014}, has a similar complex gating architecture, but requires less memory. The standard GRU architecture is shown in Fig. \ref{fig:gru}.

Because seq2seq summarization involves potentially many long-range temporal dependencies, our model applies temporal hierarchy to the GRU. We apply a timescale constant at the end of a GRU, essentially adding another constant gating unit that modulates the mixture of past and current hidden states. The reset gate $r_t$, update gate $z_t$, and the candidate activation $u_t$ are computed similarly to that of the original GRU as shown in Eq.(\ref{eqn:gru}).

\begin{figure}
    \centering
    \includegraphics[width=1\columnwidth]{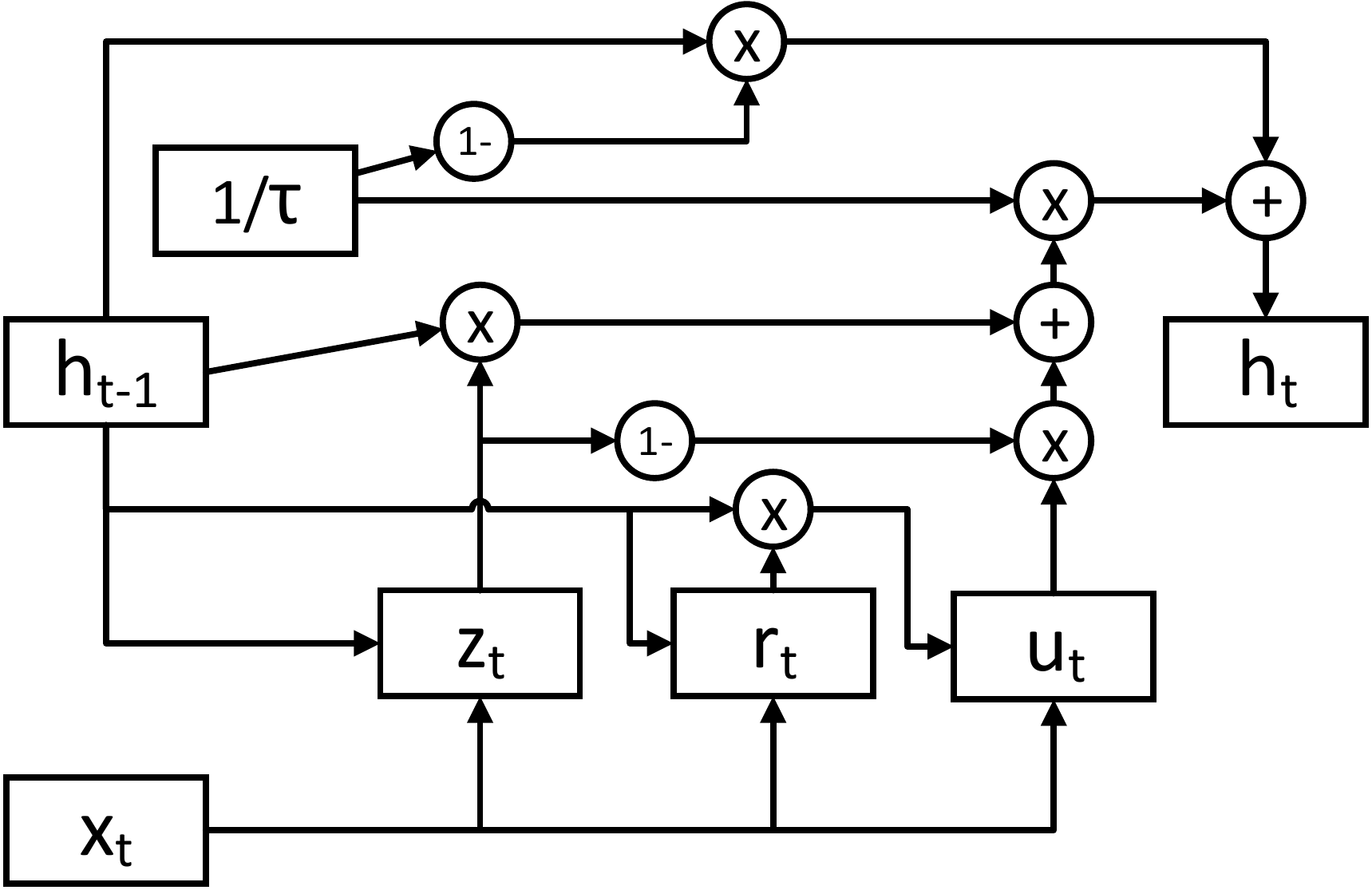}
    \caption{Proposed multiple timescale gated recurrent unit.}
    \label{fig:mtgru}
    \vspace{-2mm}
\end{figure}

\begin{equation}
\vspace{-3mm}
\label{eqn:gru}
\begin{aligned}
    r_t = \sigma( W_{xr} x_t + W_{hr} h_{t-1}) \\
    z_t = \sigma( W_{xz} x_t + W_{hz} h_{t-1}) \\
    u_t = \tanh( W_{xu} x_t + W_{hu} (  r_t \odot h_{t-1}))
\end{aligned}
\vspace{-5mm}
\end{equation}

\begin{equation}
\label{eqn:mtgru}
\begin{aligned}
     h_t = ((1 - z_t) h_{t-1} + z_t u_t)\frac{1}{\tau}+(1-\frac{1}{\tau})h_{t-1}
\end{aligned}
\end{equation}

The time constant $\tau$ added to the activation $h_t$ of the MTGRU is shown in Eq.(\ref{eqn:mtgru}). $\tau$ is used to control the timescale of each GRU cell. Larger $\tau$ meaning slower cell outputs but it makes the cell focus on the slow features of a dynamic sequence input. The proposed MTGRU model is illustrated in Fig. \ref{fig:mtgru}. The conventional GRU will be a special case of MTGRU where $\tau=1$, where no attempt is made to organize layers into different timescales.

\begin{equation}
\label{eqn:bpmtgru}
\begin{aligned}
     \frac{\delta E}{\delta h_{t-1}} =\frac{1}{\tau} [\frac{\delta E}{\delta h_t}\odot(u_t - h_{t-1})\odot\sigma'(z_t)W_{zh}]\\
     +\frac{1}{\tau} [((\frac{\delta E}{\delta h_t}\odot z_t\odot \tanh'(u_t))W_{uh})\odot r_t]\\
     +\frac{1}{\tau} [(((\frac{\delta E}{\delta h_t}\odot z_t\odot \tanh'(u_t))W_{uh})\\
     \odot\sigma'(r_t)\odot h_{t-1})W_{rh}]\\
     + \frac{1}{\tau} [\frac{\delta E}{\delta h_t}\odot (1-z_t)] + (1-\frac{1}{\tau})\frac{\delta E}{\delta h_t}
\end{aligned}
\end{equation}

Eq. (\ref{eqn:bpmtgru}) shows the learning algorithm derived for the MTGRU according to the defined forward process and the back propagation through time rules. $\frac{\delta E}{\delta h_{t-1}}$ is the error of the cell outputs at time $t-1$ and $\frac{\delta E}{\delta h_t}$ is the current gradient of the cell outputs. Different timescale constants are set for each layer where larger $\tau$ means slower context units and $\tau=1$ defines the default or the input timescale. 
Based on our hypothesis that later layers should learn features that operate over slower timescales, we set larger $\tau$ as we go up the layers. 

In this application, the question is whether the word sequences being analyzed by the RNN possess information that operates over different temporal hierarchies, as they do in the case of the continuous audio signals received by the human auditory system. We hypothesize that they do, and that word level, clause level, and sentence level compositionalities are strong candidates. In this light, the multiple timescale modification functions as a way to explicitly guide each layer of the neural network to facilitate the learning of features operating over increasingly slower timescales, corresponding to subsequent levels in the compositional hierarchy. 

\subsection{Summarization}

\begin{figure}
    \centering
    \includegraphics[width=1\columnwidth]{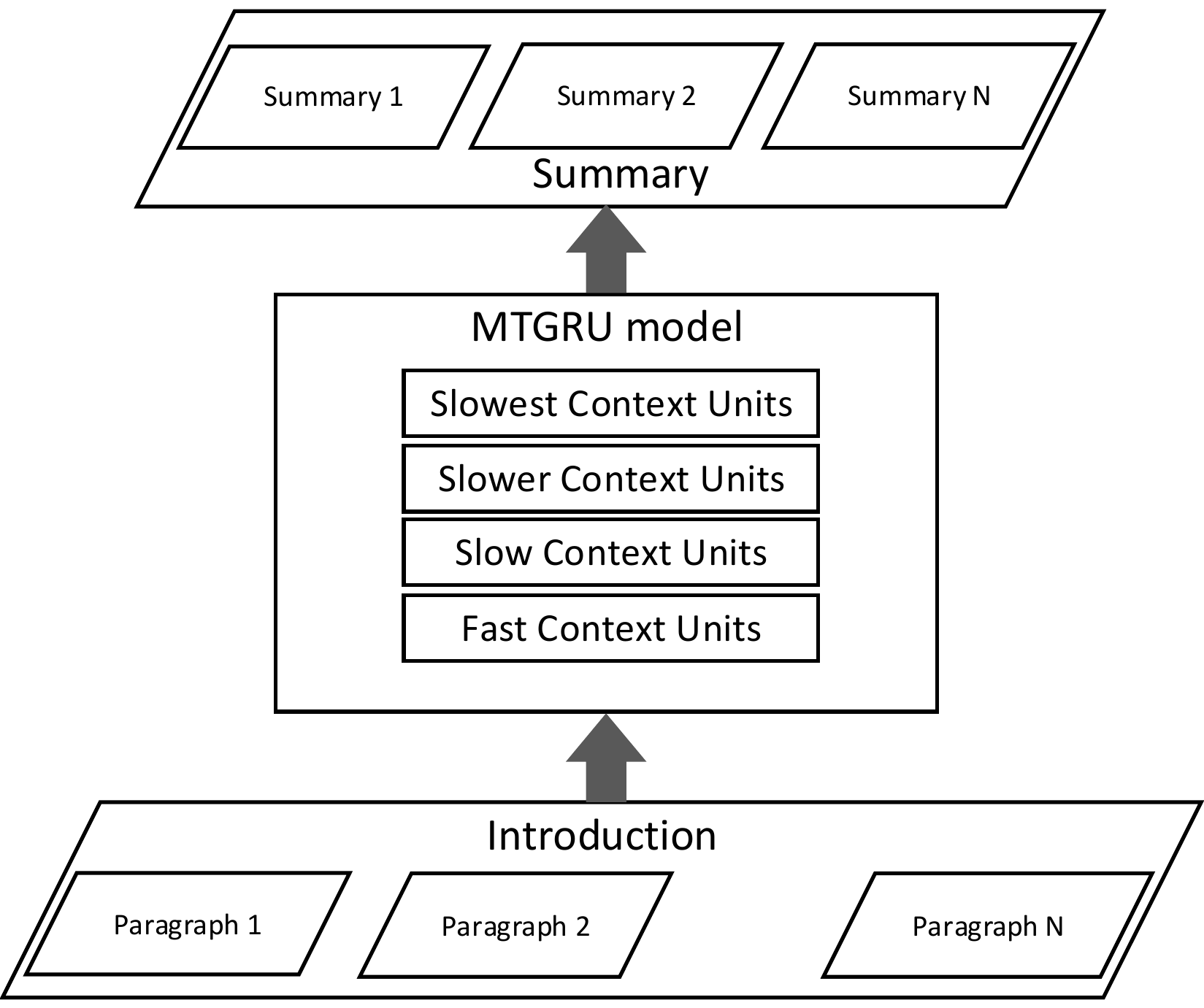}
    \caption{Paragraph level approach to summarization.}
    \label{fig:model}
    \vspace{-3mm}
\end{figure}

To apply our newly proposed multiple timescale model to summarization, we build a new dataset of academic articles. We collect LaTeX source files of articles in the CS.\{CL,CV,LG,NE\} domains from the arXiv preprint server, extracting their Introductions and Abstracts. We decompose the Introduction into paragraphs, and pair each paragraph with its most salient sentence as the target summary. These target summaries are generated using the widely adopted TF-IDF scoring. Fig. \ref{fig:model} shows the structure of our summarization model.

Our dataset contains rich compositionality and longer text sequences, increasing the complexity of the summarization problem. The temporal hierarchy function has the biggest impact when complex compositional hierarchies exist in the input data. Hence, the multiple timescale concept will play a bigger role in our context compared to previous summarization tasks such as \newcite{Namas:2015}.

The model using MTGRU is trained using these paragraphs and their targets. The generated summaries of each Introduction is evaluated using the Abstracts of the collected articles. We chose the Abstracts as gold summaries, because they usually contain important discourse structures such as goal, related works, methods, and results, making them good baseline summaries. To test the effectiveness of the proposed method, we compare it with the conventional RNN encoder-decoder in terms of training speed and performance.

\section{Experiments and Results}
\label{sec:results}
\begin{table}
\vspace{-0.1in}
\begin{center}
{\small
\begin{tabular}{|l|c|c|c|}
\hline
{\bf RNN Type}  & {\bf Layers}  & {\bf Hidden Units}  \\
\hline
\hline
GRU   &4& 1792     \\
MTGRU   &4 & 1792       \\
\hline
\end{tabular}
}
\end{center}
\vspace{-0.1in}
\caption{{\small Network Parameters for each model. }}
\label{tab:parameters}
\vspace{-0.1in}
\end{table}

We trained two seq2seq models, the first model using the conventional GRU in the RNN encoder decoder, and the second model using the newly proposed MTGRU. Both models are trained using the same hyperparamenter settings with the optimal configuration which fits our existing hardware capability.

\begin{table}
\vspace{-0.1in}
\begin{center}
{\small
\scalebox{0.95}{
\begin{tabular}{|l|c|c|c|}
\hline
{\bf Steps}&{\bf RNN}  & {\bf Train Perplexity}  & {\bf Test Perplexity}  \\
\hline
\hline
74750 &GRU&6.8&  29.72    \\
74750 &MTGRU& 5.87& 18.53      \\
\hline
\end{tabular}
}
}
\end{center}
\vspace{-0.1in}
\caption{{\small Training results of the Models. }}
\label{tab:result}
\vspace{-0.1in}
\end{table}
\begin{table}
\vspace{-0.1in}
\begin{center}
{\small
\begin{tabular}{|c|l|l|l|l|}
\hline
{\bf Evaluation Metric} & {\bf Recall}  & {\bf Precision}  & {\bf F--Score} \\
\hline
\hline
ROUGE-1 &0.48135 & 0.59030& 0.50835     \\
ROUGE-2 &0.32399 &0.39505 &0.34089     \\
ROUGE-L &0.46588 &0.57218  &0.49234     \\
\hline
\end{tabular}
}
\end{center}
\vspace{-0.1in}
\caption{{\small ROUGE scores of GRU Model}}
\label{tab:rougescores}
\vspace{-0.1in}
\end{table}

\begin{figure}
 \vspace{-2mm}
    \centering
    \includegraphics[width=1\columnwidth]{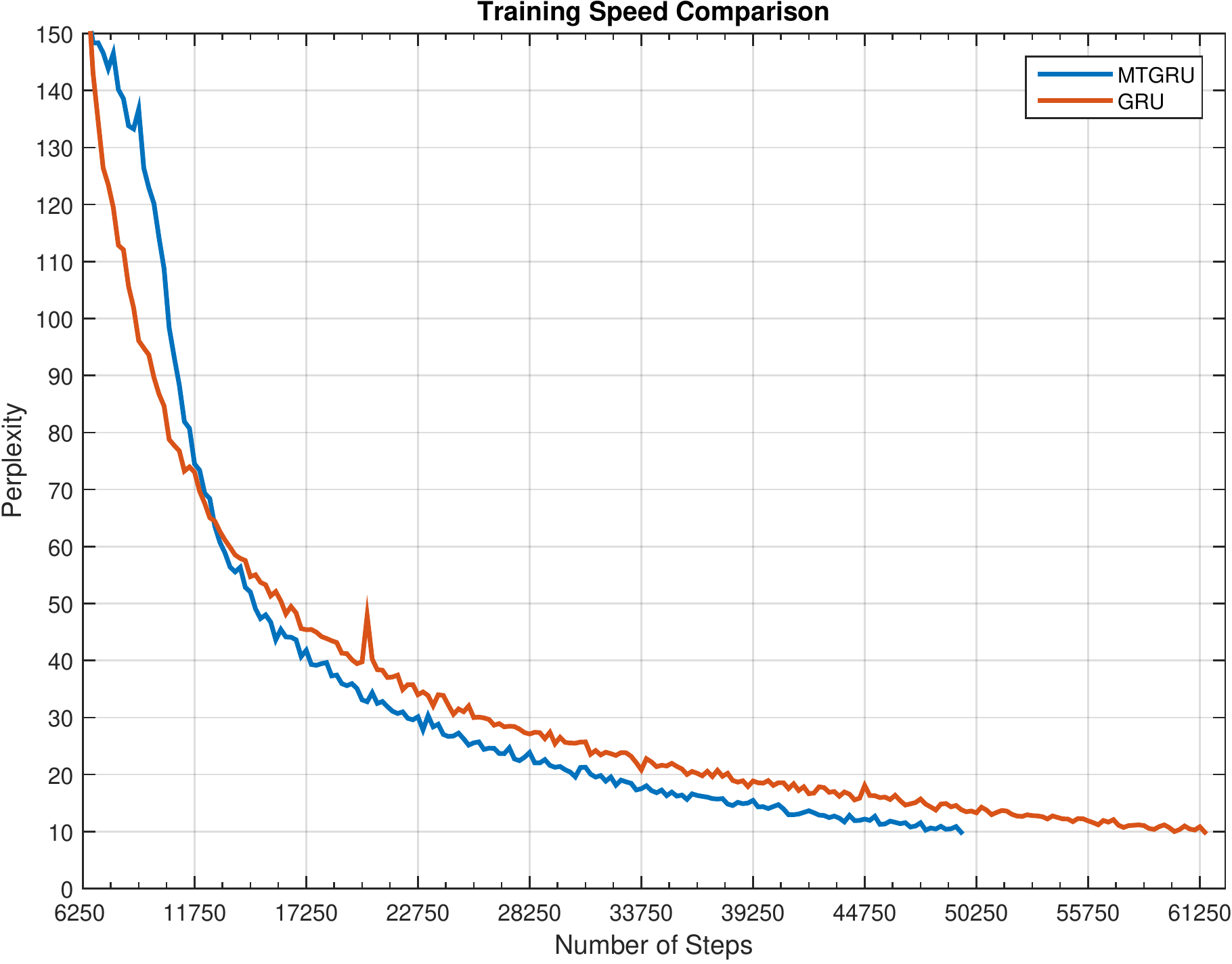}
    \caption{Comparison of Training Speed between GRU and MTGRU.}
   \label{fig:compare}
    \vspace{-1mm}
\end{figure}

Following \newcite{S2S_Ilya:2014}, the inputs are divided into multiple buckets. Both GRU and MTGRU models consist of 4 layers and 1792 hidden units. As our models take longer input and target sequence sizes, the hidden units size and number of layers are limited. An embedding size of 512 was used for both networks. The timescale constant $\tau$ for each layer is set to ${1, 1.25, 1.5, 1.7}$, respectively. The models are trained on 110k text-summary pairs. The source text are the paragraphs extracted from the introduction of academic articles and the targets are the most salient sentence extracted from the paragraphs using TF-IDF scores. For comparison of the training speed of the models, Fig. \ref{fig:compare} shows the plot of the training curve until the train perplexity reaches 9.5. Both of the models are trained using 2 Nvidia Ge-Force GTX Titan X GPUs which takes roughly 4 days and 3 days respectively.  During test, greedy decoding was used to generate the most likely output given a source Introduction.

\begin{table}
\vspace{0in}
\begin{center}
{\small
\begin{tabular}{|c|l|l|l|l|}
\hline
{\bf Evaluation Metric} & {\bf Recall}  & {\bf Precision}  & {\bf F--Score} \\
\hline
\hline
ROUGE-1 &0.50901 &0.61571 &0.53870    \\
ROUGE-2 &0.34148 &0.40824 &0.35925       \\
ROUGE-L &0.49406 &0.59830 &0.52318      \\
\hline
\end{tabular}
}
\end{center}
\vspace{-0.1in}
\caption{{\small ROUGE scores of MTGRU Model}}
\label{tab:rougescores2}
\vspace{-0.1in}
\end{table}

\begin{figure}
  \vspace{-3mm}
    \centering
    \includegraphics[width=1\columnwidth]{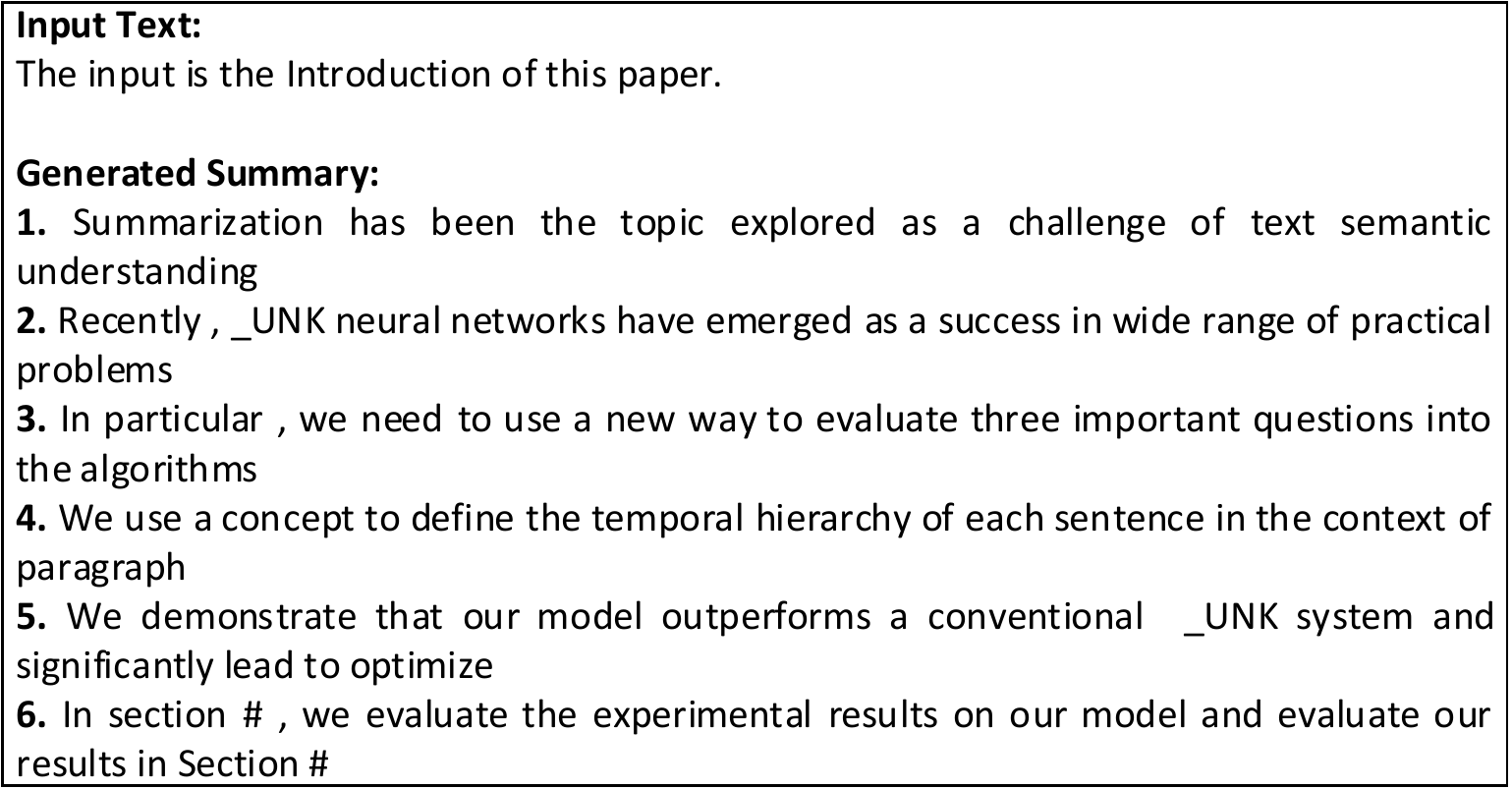}
    \caption{An example of the generated summary with MTGRU.}
   \label{fig:summary2}
    \vspace{-1mm}
\end{figure}

For evaluation, we adopt the Recall-Oriented Understudy for Gisting Evaluation (ROUGE) metrics \cite{lin2004rouge} proposed by \newcite{lin2003automatic}. ROUGE is a recall-oriented measure to score system summaries which is proven to have a strong correlation with human evaluations. It measures the \textit{n}-gram recall between the candidate summary and gold summaries. In this work, we only have one gold summary which is the Abstract of an article, thus the ROUGE score is calculated as given in \newcite{LiHRNN:2015}. ROUGE-1, ROUGE-2 and ROUGE-L are used to report the performance of the models. For the performance evaluation, both the models are trained up to 74750 steps where the training perplexity of GRU and MTGRU are shown in Table \ref{tab:result}. This step was chosen as the early stopping point as at this step we get the lowest test perplexity of the GRU model. The ROUGE scores calculated using these trained networks are shown in Table \ref{tab:rougescores} and Table \ref{tab:rougescores2} for the GRU and MTGRU models, respectively. A sample summary generated by the MTGRU model is shown in Fig. \ref{fig:summary2}. 

\begin{figure}
  \vspace{-2mm}
    \centering
    \includegraphics[width=1\columnwidth]{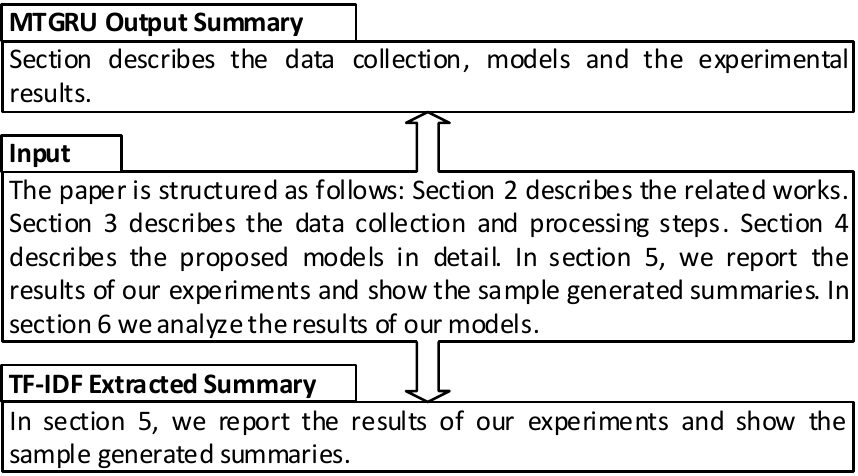}
    \caption{An example of the output summary vs the extracted targets}
   \label{fig:AbsVsExt}
    \vspace{-3mm}
\end{figure}

\section{Discussion and Future Work}
\label{sec:discussion}

The ROUGE scores obtained for the summarization model using GRU and MTGRU show that the multiple timescale concept improves the performance of the conventional seq2seq model without the presence of highly complex architectural hierarchies. Another major advantage is the increase in training speed by as much as 1 epoch. Moreover, the sample summary shown in Fig. \ref{fig:summary2} demonstrates that the model has successfully generalized on the difficult task of summarizing a large paragraph into a one line salient summary. 

\begin{figure}
 \vspace{-2mm}
    \includegraphics[width=1\columnwidth]{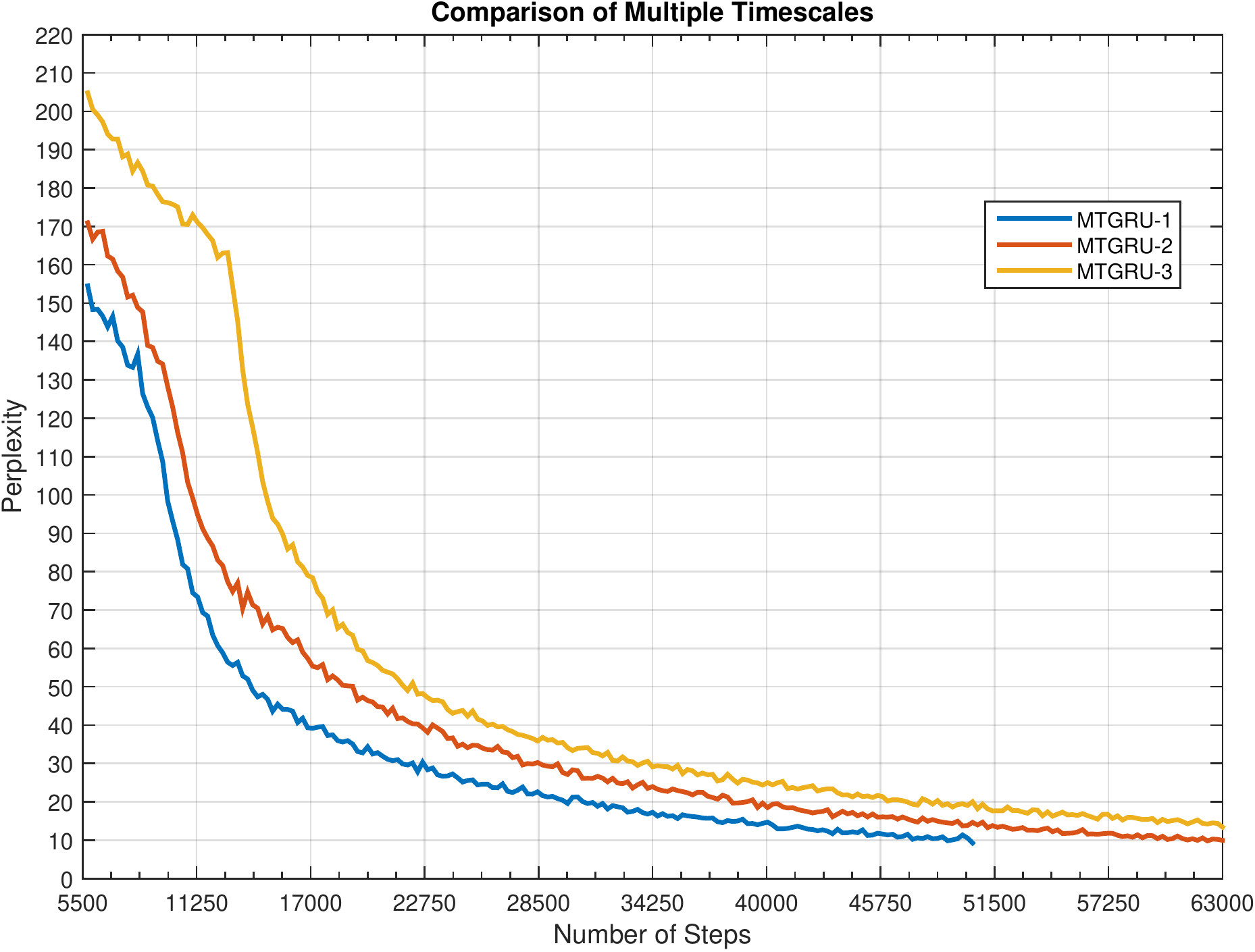}
    \caption{Comparison of Training performance between multiple time constants.}
   \label{fig:compare1}
    \vspace{-1mm}
\end{figure}

In setting the $\tau$ timescale parameters, we follow \cite{MTRNN:2008} . We gradually increase $\tau$ as we go up the layers such that higher layers have slower context units. Moreover, we experiment with multiple settings of $\tau$ and compare the training performance, as shown in Fig. \ref{fig:compare1}. The $\tau$ of MTRGU-2 and MTRGU-3 are set as \{1, 1.42, 2, 2.5\} and \{1, 1, 1.25, 1.25\}, respectively. MTGRU-1 is the final model adopted in our experiment described in the previous section. MTGRU-2 has comparatively slower context layers and MTGRU-3 has two fast and two slow context layers. As shown in the comparison, the training performance of MTRGU-1 is superior to the remaining two, which justifies our selection of the timescale settings. 

The results of our experiment provide evidence that an organizational process akin to functional differentiation occurs in the RNN in language tasks. The MTGRU is able to train faster than the conventional GRU by as much as 1 epoch. We believe that the MTRGU expedites a type of functional differentiation process that is already ocurring in the RNN, by explicitly guiding the layers into multiple timescales, where otherwise this temporal hierarchical organization occurs more gradually. 

In Fig. \ref{fig:AbsVsExt}, we show the comparison of a generated summary of the input paragraph to an extracted summary. As seen in the example, our model has successfully extracted the key information from multiple sentences and reproduces it into a single line summary. While the system was trained only on the extractive summary, the abstraction of the entire paragraph is possible because of the generalization capability of our model. The seq2seq objective maximizes the joint probability of the target sequence conditioned on the source sequence. When a summarization model is trained on source-extracted salient sentence target pairs, the objective can be viewed as consisting of two subgoals: One is to correctly perform saliency finding (importance extraction) in order to identify the most salient content, and the other is to generate the precise order of the sentence target. In fact, during training, we observe that the optimization of the first subgoal is achieved before the second subgoal. The second subgoal is fully achieved only when overfitting occurs on the training set. The generalization capability of the model is attributable to the fact that the model is expected to learn multiple points of saliency per given paragraph input (not only a single salient section corresponding to a single sentence) as many training examples are seen. This explains how the results such as those in Fig. \ref{fig:AbsVsExt} can be obtained from this model. 

We believe our work has some meaningful implications for seq2seq abstractive summarization going forward. First, our results confirm that it is possible to train an encoder-decoder model to perform saliency identification, without the need to refer to an external corpus at test time. This has already been shown, implicitly, in previous works such as \newcite{Namas:2015,Nallapati2016}, but is made explicit in our work due to our choice of data consisting of paragraph-salient sentence pairs. Secondly, our results indicate that probabilistic language models can solve the task of novel word generation in the summarization setting, meeting a key criteria of abstractive summarization. \newcite{bengio2003neural} originally demonstrated that probabilistic language models can achieve much better generalization over similar words. This is due to the fact that the probability function is a smooth function of the word embedding vectors. Since similar words are trained to have similar embedding vectors, a small change in the features induces a small change in the predicted probability. This makes a strong case for RNN language models as the best available solution for abstractive summarization, where it is necessary to generate novel sentences. For example, in Fig. \ref{fig:summary2}, the first summary shows that our model generates the word ``explored" which is not present in the paper. Furthermore, our results suggest that if given abstractive targets, the same model could train a fully abstractive summarization system. 

In the future, we hope to explore the organizational effect of the MTGRU in different tasks where temporal hierarchies can arise, as well as investigating ways to effectively optimize the timescale constant. Finally, we will work to move towards a fully abstractive end-to-end summarization system of multi-paragraph text by utilizing a more abstractive target which can potentially be generated with the help of the Abstract from the articles. 

\section{Conclusion}
In this paper, we have demonstrated the capability of the MTGRU in the multi-paragraph text summarization task. Our model fulfills a fundamental requirement of abstractive summarization, deep semantic understanding of text and importance identification. The method draws from a well-researched phenomenon in the human brain and can be implemented without any hierarchical architectural complexity or additional memory requirements during training. Although we show its application to the task of capturing compositional hierarchies in text summarization only, MTGRU also shows the ability to enhance the learning speed thereby reducing training time significantly. In the future, we hope to extend our work to a fully abstractive end-to-end summarization system of multi-paragraph text. 

\section*{Acknowledgment}
This research was supported by Basic Science Research Program through the National Research Foundation of Korea(NRF) funded by the Ministry of Science, ICT and future Planning(2013R1A2A2A01068687) (50\%), and by the Industrial Strategic Technology Development Program (10044009) funded by the Ministry of Trade, Industry and Energy (MOTIE, Korea) (50\%).

\bibliography{acl2016}
\bibliographystyle{acl2016}

\end{document}